%% file: main.tex
\documentclass[10pt,twocolumn,letterpaper]{article}

\usepackage[pagenumbers]{cvpr} %

\usepackage{siunitx}
\usepackage{tcolorbox}
\usepackage{adjustbox}
\usepackage{tikz}
\usetikzlibrary{positioning}
\usepackage{colortbl} %
\usepackage{xcolor}

\input{preamble}

\definecolor{cvprblue}{rgb}{0.21,0.49,0.74}

\usepackage[pagebackref,breaklinks,colorlinks,allcolors=cvprblue]{hyperref}

\def\paperID{9483} %
\def\confName{CVPR}
\def\confYear{2025}

\title{LLaMA-Mesh: Unifying 3D Mesh Generation with Language Models}

\author{
\normalsize{Zhengyi Wang\textsuperscript{1 *} \hspace{0.02\textwidth} Jonathan Lorraine\textsuperscript{2} \hspace{0.02\textwidth} Yikai Wang\textsuperscript{1} \hspace{0.02\textwidth} Hang Su\textsuperscript{1} \hspace{0.02\textwidth} Jun Zhu\textsuperscript{1}
\hspace{0.02\textwidth} Sanja Fidler\textsuperscript{2} \hspace{0.02\textwidth} Xiaohui Zeng\textsuperscript{2}}\\
\normalsize{{Tsinghua University\textsuperscript{1} \hspace{0.02\textwidth} NVIDIA\textsuperscript{2}}}\\
\normalsize{\url{https://research.nvidia.com/labs/toronto-ai/LLaMA-Mesh}}
}

\begin{document}

\twocolumn[{
    \renewcommand\twocolumn[1][]{#1}
    \maketitle
    \begin{center}
    \vspace{-0.045\textheight}
    \centering
        \includegraphics[width=1.02\linewidth,trim={0.0cm 0 0 0},clip]{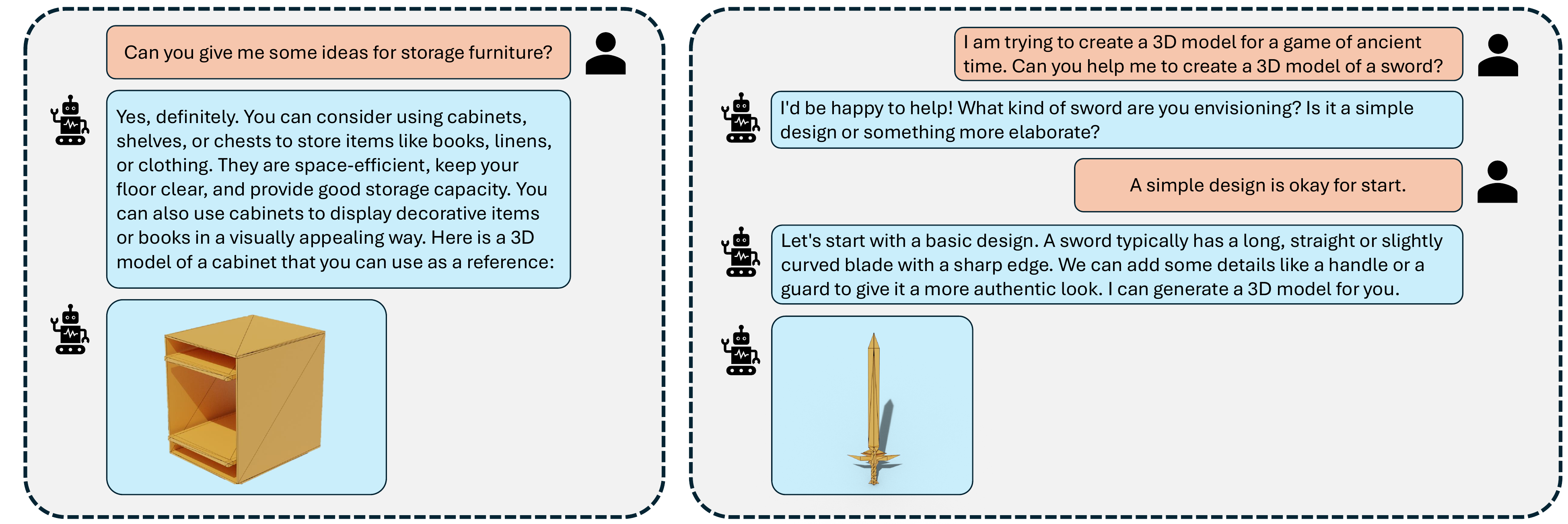}
    \vspace{-0.03\textheight}
    \captionof{figure}[Short caption]{
    An illustration of our method, \ours, which enables the generation of 3D meshes from human instructions via a conversational interface. Users provide textual prompts, and the model responds with both text and 3D mesh outputs, facilitating interactive 3D content creation. \ours allows large language models to generate and interpret 3D meshes from text directly, seamlessly unifying language and 3D modalities within a single model.
    }
    \label{fig:teaser-top}
    \vspace{-0.005\textheight}
    \end{center}
}]

\begin{abstract}

    \vspace{-.01\textheight}
    This work explores expanding the capabilities of large language models (LLMs) pretrained on text to generate 3D meshes within a unified model. This offers key advantages of (1) leveraging spatial knowledge already embedded in LLMs, derived from textual sources like 3D tutorials, and (2) enabling conversational 3D generation and mesh understanding. A primary challenge is effectively tokenizing 3D mesh data into discrete tokens that LLMs can process seamlessly. To address this, we introduce \ours, a novel approach that represents the vertex coordinates and face definitions of 3D meshes as plain text, allowing direct integration with LLMs without expanding the vocabulary. We construct a supervised fine-tuning (SFT) dataset enabling pretrained LLMs to (1) generate 3D meshes from text prompts, (2) produce interleaved text and 3D mesh outputs as required, and (3) understand and interpret 3D meshes. Our work is the first to demonstrate that LLMs can be fine-tuned to acquire complex spatial knowledge for 3D mesh generation in a text-based format, effectively unifying the 3D and text modalities. \ours achieves mesh generation quality on par with models trained from scratch while maintaining strong text generation performance.
    \begingroup\renewcommand{\thefootnote}{}\footnotemark\endgroup
    \renewcommand{\thefootnote}{\fnsymbol{footnote}}
    \footnotetext[1]{Work completed during NVIDIA internship.}
    \vspace{-0.015\textheight}
\end{abstract}

    \input{figs/tex/pipeline}

\section{Introduction}\label{sec:intro}

    Large Language Models (LLMs)~\cite{vaswani2023attention,brown2020language} have demonstrated remarkable capabilities in understanding and generating human-like text, achieving success in applications such as conversational agents, code generation, and visual content reasoning~\cite{achiam2023gpt,dubey2024llama,liu2023llava}.
    Despite these advances, their generative abilities have primarily been limited to the textual content, restricting their utility for broader tasks.

Our work seeks to extend LLMs into a new modality—3D mesh generation—unlocking significant potential for fields like computer graphics, engineering, robotics, and virtual/augmented reality. By enabling LLMs to generate 3D meshes from textual descriptions, we unify language understanding with 3D content creation, expanding the functional scope of LLMs. This approach paves the way for more intuitive and efficient workflows in 3D content creation driven by language-based instructions.

    However, integrating a new modality into an LLM is challenging, particularly in the tokenization process for processing the new modality. To the best of our knowledge, there have been no attempts to unify 3D mesh and text generation in a single framework. Some studies explored unifying image and text generation. Among these works~\cite{lu2023chameleon,wang2024emu3}, a common approach is to train a new tokenizer such as a vector-quantized variational autoencoder (VQ-VAE)~\cite{oord2018vqvae,esser2021taming} to encode the new modality into discrete tokens, which are used in training. However, this requires vocabulary expansion, increasing the adaptation's learning cost. 
    Additionally, this method introduces information loss during the auto-encoding process.

    To tackle these challenges, we introduce \ours, a novel framework that enables large language models (LLMs) to generate 3D meshes by representing them as plain text. Our approach uses the OBJ file format, a widely adopted text-based standard for 3D models comprising vertex coordinates and face definitions, as shown in Figure~\ref{fig:mesh_representation}. By treating these numerical values as a sequence of text, 
    we convert 3D meshes into a format that LLMs can process directly, avoiding modifications to the tokenizer or the vocabulary, thus minimizing additional training overhead. This design capitalizes on the extensive knowledge embedded in pretrained LLMs. Figure~\ref{fig:dialog_pretrain_LLM} shows pretrained LLMs demonstrate a native ability to represent 3D structures in text — a capability our framework harnesses.
    
    We construct a supervised fine-tuning (SFT) dataset that includes text-3D pairs and interleaved text-3D dialogues. We fine-tune a pretrained LLaMA-3.1-8B-Instruct~\cite{dubey2024llama} model on our curated dataset. 
    We find that LLMs can acquire complex spatial knowledge by learning the numerical values of meshes in textual format.
    After fine-tuning, our model demonstrates the ability to (1) generate 3D meshes given text prompts, (2) produce interleaved outputs of text and 3D meshes in a conversational setup, and (3) describe meshes in natural language. %

    \ours is the first successful effort to empower an LLM to generate 3D content with language, unifying the 3D and text modalities in a single large model. It achieves mesh generation quality comparable to models trained from scratch while maintaining strong text generation abilities.

    \input{figs/tex/gallery_mesh_gen}

\section{Related Work}
    We overview approaches to make LLMs multi-modal in Section~\ref{sec:related_multi_modal_llm}, 3D object generation more broadly in Section~\ref{sec:related_3dgen}, and then specifically generating meshes auto-regressively in Section~\ref{sec:related_meshgen}.

    \subsection{Enabling LLMs to be Multi-Modal}\label{sec:related_multi_modal_llm}
        Extending LLMs to process and generate multiple modalities, such as vision and language, in a unified model is an active research area.
        Existing works allow an LLM to understand visual input for multimodal interactions~\citep{li2022blip,li2023blip2,chen2024internvl,lin2024vila,alayrac2022flamingo,zhang2023internlm,liu2024improved,ye2023mplug, liao2024reasoning, liu2024visual, bai2023qwen} or unify image and text generation~\citep{xie2024show,zhou2024transfusion,wu2024vila,yu2024spae, wang2024emu3, lu2023chameleon} with a new visual tokenizer.
        Instead, we avoid modifying tokenization and focus on 3D by simply outputting (the text of) an OBJ file.
        
        Closely related are works where LLMs wield tools to generate 3D scenes~\citep{yang2024holodeck, zhang2024scene} by generating layouts to compose predefined objects.
        However, these methods do not enable LLMs to produce 3D meshes directly.
        To the best of our knowledge, we are the first to allow LLMs to directly generate 3D meshes as text instead of just wielding 3D object generation tools.
    
    \subsection{3D Object Generation}\label{sec:related_3dgen}
        DreamFusion~\cite{poole2022dreamfusion}, Magic3D~\cite{lin2023magic3d}, ProlificDreamer~\cite{wang2023prolificdreamer} and many other methods~\cite{chen2023fantasia3d,zhu2023hifa,chen2023it3d,tang2023dreamgaussian,chen2023gsgen,liang2023luciddreamer,sun2023dreamcraft3d, lorraine2023att3d, yu2023csd,liu2023sherpa3d,kim2023collaborative,wang2023animatabledreamer, xie2024latte3d} use score-distillation to generate 3D objects from pretrained large-scale text-to-image diffusion model~\cite{saharia2022photorealistic,rombach2022high}. Feed-forward methods including LRM~\cite{hong2023lrm,li2023instant3d,xu2023dmv3d,wang2023pflrm}, CRM~\cite{wang2024crm}, InstantMesh~\cite{xu2024instantmesh}, and other methods~\cite{zou2023triplane,tang2024lgm,liu2023one,liu2023onepp,zhang2024clay,li2024craftsman,wu2024direct3d} generate 3D objects without test-time optimization. However, the above methods typically treat 3D objects as numerical fields and extract meshes using marching cubes or their variants~\cite{shen2021dmtet,shen2023flexible}, which do not easily allow representation as discrete tokens.

    \subsection{Auto-Regressive Mesh Generation}\label{sec:related_meshgen}
        Methods such as PolyGen~\cite{nash2020polygen}, MeshGPT~\cite{siddiqui2024meshgpt}, MeshXL~\cite{chen2024meshxl}, instead model a 3D object as a discrete sequence of tokenized coordinates and use an auto-regressive transformer to generate an object with artist-created topology. MeshAnything~\cite{chen2024meshanything,chen2024meshanythingv2}, PivotMesh~\cite{weng2024pivotmesh} and EdgeRunner~\cite{tang2024edgerunner} take point cloud as input condition for better control. These works also treat meshes as discrete tokens generated using auto-regressive transformers, but they are trained from scratch and lack language capabilities.

\section{Method}\label{sec:method}
    We now introduce \ours.
    First, in Section~\ref{sec:method_meshRepr}, we explain why and how we represent 3D meshes as plain text for easy processing by LLMs.
    Then, in Section~\ref{sec:method_pretrainedModels}, we detail the pretrained LLaMA model~\cite{dubey2024llama}, an effective initialization of \ours.
    Finally, Section~\ref{sec:method_finetuning} describes how we create a 3D dialog SFT dataset to give LLMs 3D generation capabilities through fine-tuning. Our model structure is illustrated in Figure~\ref{fig:pipeline}.

    \input{figs/tex/mesh_representation}
    \subsection{3D Representation}\label{sec:method_meshRepr}

        To enable large language models (LLMs) to generate 3D meshes directly, a key challenge lies in tokenizing this new modality so that the LLM can process it effectively. We observe that pretrained LLMs can generate 3D objects in the OBJ file format—a simple and widely used plain text format—in a zero-shot manner, as illustrated in Figure~\ref{fig:dialog_pretrain_LLM}. Although these generated shapes are simple and not immediately usable, they demonstrate that some 3D knowledge in OBJ format is inherently encoded in the LLMs. Additionally, since OBJ files describe 3D geometry in a plain text format, they are ideal candidates for integration with LLMs without requiring modifications to the tokenizer or vocabulary. These insights motivate us to represent 3D objects using the OBJ file format.

        An OBJ file consists of a list of vertex coordinates and face definitions:
        \textbf{Vertices (v):} Each line starting with the letter \texttt{v} defines a vertex in 3D space with its \texttt{x}, \texttt{y}, and \texttt{z} coordinates, e.g., \texttt{v 0.123 0.234 0.345}.
        \textbf{Faces (f):} Each line starting with the letter \texttt{f} defines a face by listing vertex indices that form a polygon (typically a triangle or quadrilateral), e.g., \texttt{f 1 2 3}. By treating these numerical values as plain text, we convert the 3D mesh into a sequential text format that LLMs process natively. Figure~\ref{fig:mesh_representation} shows an example of a simple OBJ file and its corresponding 3D object rendering. Note that OBJ files from the internet may vary slightly in format. We adopt a standard that is widely used and straightforward.
        
        Note that 3D mesh coordinates are typically stored as floating-point numbers. Using floating-point numbers for vertex coordinates directly leads to long token sequences, exceeding most LLM's context length limitations and increasing the computational cost. To address this, we quantize the vertex coordinates into a fixed number of bins (\num{64} per axis in our case). We scale the mesh to the range $[0,64]$ and quantize the coordinates to the nearest integer. Figure~\ref{fig:tokenizer2} shows our quantization, which slightly reduces the coordinates' precision. However, it significantly decreases the token count, 
        making it feasible for LLMs to handle longer sequences without sacrificing geometric fidelity.

    \input{figs/tex/tokenizer2}

    \subsection{Pretrained Models}\label{sec:method_pretrainedModels}
        Pretrained LLMs, like LLaMA~\cite{dubey2024llama} variants, are natural candidates for generating text for meshes as they are (a) strong tools for modeling arbitrary sequences and (b) may have encountered similar data during pre-training.
        Specifically, we use LLaMA3.1-8B-Instruct~\cite{dubey2024llama} as our base model.
        
        This model is chosen for its balance between performance and computational efficiency. It has been instruction-tuned to follow prompts and generate coherent responses, which is advantageous for our application, where the model needs to interpret text prompts and generate corresponding 3D meshes.
        Notably, the model can generate simple (but not perfect) OBJ files without fine-tuning, as shown in Figure~\ref{fig:dialog_pretrain_LLM}, likely because there are publicly available examples, e.g., on GitHub. 
        
        Despite its strengths, the pretrained LLaMA model performs poorly on mesh generation tasks without fine-tuning, underscoring the need for fine-tuning the model on a specialized dataset that includes mesh representations in plain text.
        By fine-tuning LLaMA on our curated dataset of text-mesh pairs, we enable the model to learn the patterns and semantics of the OBJ format, allowing it to generate valid 3D meshes directly from textual descriptions.

        \input{figs/tex/dialog_pretrain_LLM}

    \input{figs/tex/more_dialog_results} 
    \subsection{3D-task Finetuning}
        \label{sec:method_finetuning}
        
    To equip LLMs with 3D capabilities, we construct a supervised fine-tuning (SFT) dataset for training. We use 3D meshes from Objaverse~\cite{deitke2023objaverse}, a comprehensive 3D dataset for general objects. To build the chat dataset, we employ (1) a rule-based approach and (2) LLM-based augmentation.
    
    In the rule-based approach, we design several simple patterns, such as \textit{``(user) \{obj\} What is this? (assistant) \{caption\}."} for mesh understanding, and \textit{``(user) Create a 3D model of \{caption\}. (assistant) \{obj\}."} for mesh generation. For each 3D object, we randomly select a pattern and replace placeholders with the mesh definition and caption. Although these conversations are straightforward, they provide the LLM with foundational knowledge of the correspondence between text and 3D representations.

    To enable more sophisticated conversations, we create complex text-3D dialogues. We write sample dialogues in a text-3D interleaved format and use in-context learning to prompt the pretrained LLM to generate dialogues for each 3D object based on its textual description.
    We use a combination of rule-based and LLM augmentation methods. We randomly choose from the rule-based approach and LLM-based augmentation for each mesh to construct the dialog.  Figure~\ref{fig:training_data} shows examples of our training data.
    
    To preserve the  LLM's language capabilities, we use UltraChat~\cite{ding2023ultrachat}, a general conversational dataset. Our final dataset is a mix of mesh generation, mesh understanding, and general conversation data, using the ratio in Table~\ref{tab:dataset_stat}.

\input{figs/tex/training_data}
\section{Experiments}
    We first provide implementation details in Section~\ref{sec:exp_implementation}, including dataset preparation and training.
    Next, in Section~\ref{sec:exp_results}, we include our method's results, showcasing the quality and diversity of generated meshes and the chatting ability preserved from the pretrained LLM.
    We compare \ours with baseline methods in Section~\ref{sec:exp_baselines}.%

    \subsection{Implementation Details}\label{sec:exp_implementation}
        \paragraph{Dataset preparation}
            We filter the Objaverse dataset~\citep{deitke2023objaverse} to select meshes with a maximum of $500$ faces to maintain a manageable computational complexity, resulting in $31k$ total meshes. Each is converted to the OBJ format, and vertex coordinates are quantized into $64$ bins to reduce the token sequence length without significantly compromising geometric detail. We use captions generated by Cap3D~\citep{luo2024cap3d} as text descriptions accompanying each mesh. To avoid overfitting, we randomly rotate the meshes with degrees from $\{0^{\circ},90^{\circ},180^{\circ},270^{\circ}\}$, resulting in around $125k$ meshes. Following prior works~\cite{weng2024pivotmesh}, we sort the vertices by $z$-$y$-$x$ coordinates from lowest to highest. We also sort faces by lowest vertex indices. If two faces have the same lowest index, we sort by the next lowest, and so on. The LLM's context length is set to $8k$ tokens.
    
    \input{figs/tex/train_loss}
        \paragraph{Training}

            The model is trained on $32$ A100 GPUs for $21k$ iterations. We conduct full parameter fine-tuning. We use the AdamW optimizer~\citep{loshchilov2017decoupled}, with a learning rate of $1e-5$, a warm-up of $30$ steps with cosine scheduling, and a global batch size of $128$. The total training time is around $3$ days. We visualize the training loss in Figure~\ref{fig:train_loss}, which shows the model converges rapidly on the new modality, indicating fast adaptation of knowledge. We do not observe loss spiking or instabilities during training.

    \input{figs/tex/dataset_stat}
    \input{figs/tex/mesh_gen_diversity}
    \subsection{Results}\label{sec:exp_results}
        \subsubsection{Mesh Generation Results}\label{sec:exp_diversity}
            Figure~\ref{fig:gallery} shows our method generates high-quality meshes. Similarly to previous auto-regressive mesh generation methods~\cite{siddiqui2024meshgpt,chen2024meshanything}, our approach produces artist-like topology, as it learns mesh topology during training.
            
            We evaluate the diversity of generated meshes by providing the same text prompt multiple times and observing the variations in the resulting meshes. Figure~\ref{fig:diversity} demonstrates the model generates a variety of unique meshes that all satisfy the prompt, highlighting our ability to produce diverse and creative outputs. This diversity is essential for applications requiring multiple design options or variations.

    \input{figs/tex/baseline_mesh_gen}
    \input{figs/tex/training_time}
        \subsubsection{Language and Conversational Abilities}\label{sec:exp_conversation}            
            \paragraph{Qualitative Results}
                After fine-tuning for mesh generation, we evaluate whether \ours{} retains its language understanding capabilities. Figures~\ref{fig:teaser-top} and~\ref{fig:more_dialog_results} show the model engages in coherent and contextually appropriate dialogues, comprehending complex instructions, asking clarifying questions, and providing detailed responses, demonstrating its language proficiency remains intact.
            \paragraph{Quantitative Results}
                Table~\ref{tab:ablation_llm_benchmark} presents quantitative results evaluating language abilities. We report the performance of our model, \ours (8B), and compare it with baseline models of various sizes: LLaMA3.1 (8B), LLaMA3.2 (3B), and LLaMA3.2 (1B). The metrics include MMLU~\cite{hendrycks2020mmlu} (5-shot), PIQA~\cite{bisk2020piqa} (0-shot), HellaSwag~\cite{zellers2019hellaswag} (0-shot), and GSM8K~\cite{cobbe2021gsm8k} (8-shot), which assess the model’s general knowledge, commonsense reasoning, and mathematical problem-solving skills. Our model, fine-tuned to generate OBJ files for 3D mesh generation, retains language understanding and reasoning capabilities comparable to the baseline models. This demonstrates that \ours successfully extends the LLM’s functionality to 3D content generation while preserving its original language capabilities.

\input{figs/tex/ablation_llm_benchmark}

    \subsection{Comparison with Existing Methods}\label{sec:exp_baselines}
        
        \paragraph{Qualitative Comparison}
            To assess our 3D generation capability, we compare with state-of-the-art methods in 3D mesh generation, specifically MeshXL~\cite{chen2024meshxl}, an auto-regressive text-to-mesh generation approach, and Unique3D~\cite{wu2024unique3d}, a 3D generation method based on multi-view image diffusion. Since Unique3D produces dense meshes, we do not visualize their topology. Additionally, because Unique3D is originally designed for image-to-3D generation, we use SDXL~\cite{podell2023sdxl} to generate an input image for Unique3D based on the text prompt. Figure~\ref{fig:baseline_mesh_gen} shows \ours generates meshes of comparable quality to existing methods when given the same text prompt, capturing fine details and complex geometries effectively. While Unique3D and MeshXL are specialized models trained exclusively for mesh generation, \ours achieves similar results while maintaining robust language understanding capabilities within a single model.        
        \paragraph{Training Efficiency and Model Size}
            Table~\ref{tab:training_time} compares training time, computational resources, and model sizes across different methods. MeshXL~\cite{chen2024meshxl} trains large transformer models entirely on mesh data, demanding substantial computational resources. In contrast, leveraging a pretrained LLM makes our fine-tuning approach considerably more efficient, reducing training computational cost.

\newpage
\section{Discussion}\label{sec:discussion}

    \subsection{Limitations}\label{sec:limitations}
        While \ours{} shows the LLM's potential for 3D mesh generation, there are several limitations to address in future work.
        Quantizing vertex coordinates into a limited number of bins can lead to a loss of geometric detail, affecting the generated meshes' fidelity.
        Also, the context length constraints the model's ability to generate highly complex or large-scale 3D structures. Currently, we only support a maximum of $500$ faces, limiting the mesh detail level.

        We observe a slight degradation in language ability after fine-tuning as in Table~\ref{tab:ablation_llm_benchmark}. We conjecture this is due to relying solely on UltraChat as our text instruction dataset. Incorporating more diverse and high-quality text instruction datasets could help preserve the language capabilities of \ours.
        Also, we only use 3D meshes from Objaverse~\cite{deitke2023objaverse} dataset for training. We believe incorporating more datasets could enrich the generated results. Additionally, we use only an 8B model due to limited computational resources; we believe that using a larger LLaMA model would further improve the results.
        
    \subsection{Conclusion}\label{sec:conclusion}
        We introduced \ours, a novel approach that unifies 3D mesh generation with large language models by representing meshes as plain text. By fine-tuning LLaMA on a 3D dialog dataset we curated, we enabled it to generate 3D meshes directly from textual prompts without expanding the vocabulary or introducing new tokenizers. Our method preserves the language understanding capabilities of the base model while extending its generative abilities to the 3D domain. Experimental results show that \ours achieves mesh generation quality comparable to specialized models trained from scratch on 3D data. This work represents a significant step toward integrating multi-modal content generation within a cohesive language model.

    \subsection{Future Work}\label{sec:future}
        Future work could explore more efficient encoding schemes for 3D data within language models, methods to handle longer context lengths, and techniques to improve the geometric precision of generated meshes. Integrating additional modalities, such as textures or physical properties, and extending the model's capabilities to handle dynamic scenes are promising directions.
        
        Integrating 3D mesh generation into LLMs opens up exciting possibilities for interactive design, where users can converse with a model to create and manipulate 3D objects in real time. Such advancements could revolutionize virtual reality, gaming, education, and manufacturing by making 3D content creation more intuitive. We envision a future where large language models are universal generative tools capable of seamlessly producing content across multiple modalities, including text, images, and 3D structures. %

    \section*{Reproducibility}
        We provide comprehensive implementation details in Section~\ref{sec:method}, including data preprocessing steps, model architecture specifications, and training procedures. All hyperparameters, training configurations, and evaluation protocols are detailed in Section~\ref{sec:exp_implementation}.

    \section*{Ethics Statements}
        Our work unifies 3D mesh generation with LLMs, and thus we may inherit ethical concerns from both language models and generative models in general. Potential risks include generating 3D content that could be misused for malicious purposes, reproducing biased or inappropriate content present in training data, and potentially impacting the livelihoods of 3D artists and designers due to automation.
    
    \section*{Acknowledgements}\label{sec:ack}
        \noindent
        We thank Jiahui Huang, Haoxiang Wang, Yiwen Chen, David Acuna and Amlan Kar for their helpful feedback.

\newpage
{
    \small
    \bibliographystyle{ieeenat_fullname}
    \bibliography{main}
}
\newpage

\end{document}

%% file: preamble.tex
\newif\ifcomments

\commentstrue
\newcommand{\ours}{\textsc{Llama-Mesh}\xspace}

%% file: figs/tex/pipeline.tex
\begin{figure*}
    \centering
    \includegraphics[width=\linewidth]{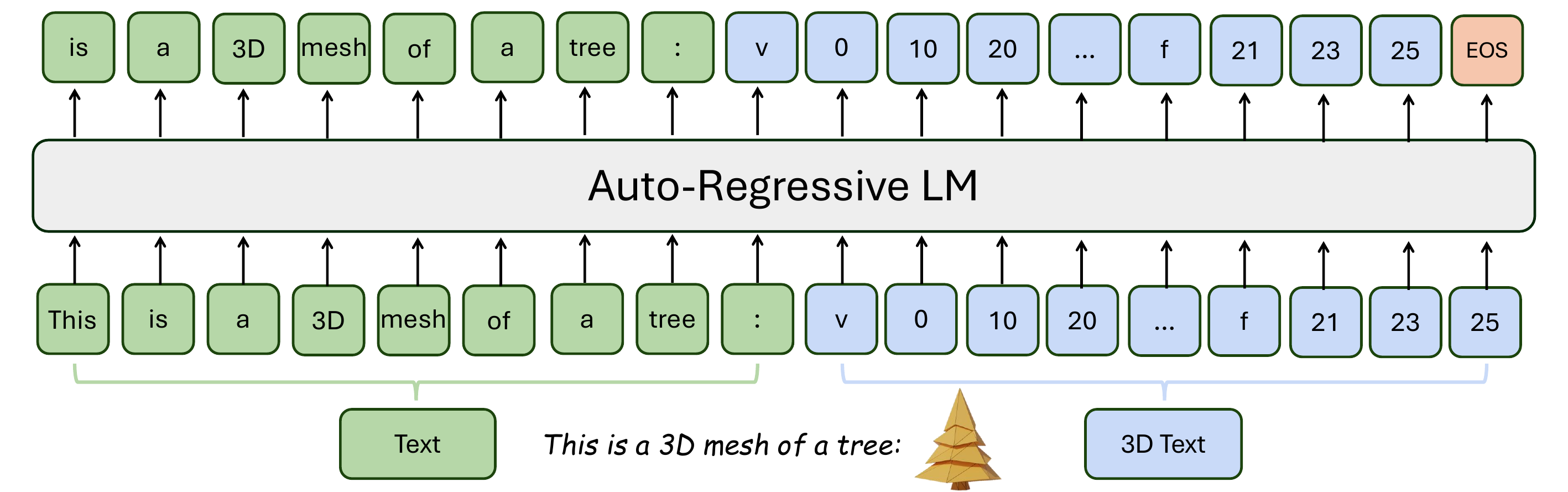}
    \vspace{-0.01\textheight}
    \caption{Overview of our method. \ours unifies text and 3D mesh in a uniform format by representing the numerical values of vertex coordinates and face definitions of a 3D mesh as plain text. Our model is trained using text and 3D interleaved data end-to-end. Therefore, with a single, unified model, we can generate both text and 3D meshes.}
    \label{fig:pipeline}
    \vspace{-0.01\textheight}
\end{figure*}

%% file: figs/tex/gallery_mesh_gen.tex
    \begin{figure*}[t]
      \centering
      \vspace{-0.03\textheight}
      \includegraphics[width=0.99\linewidth]{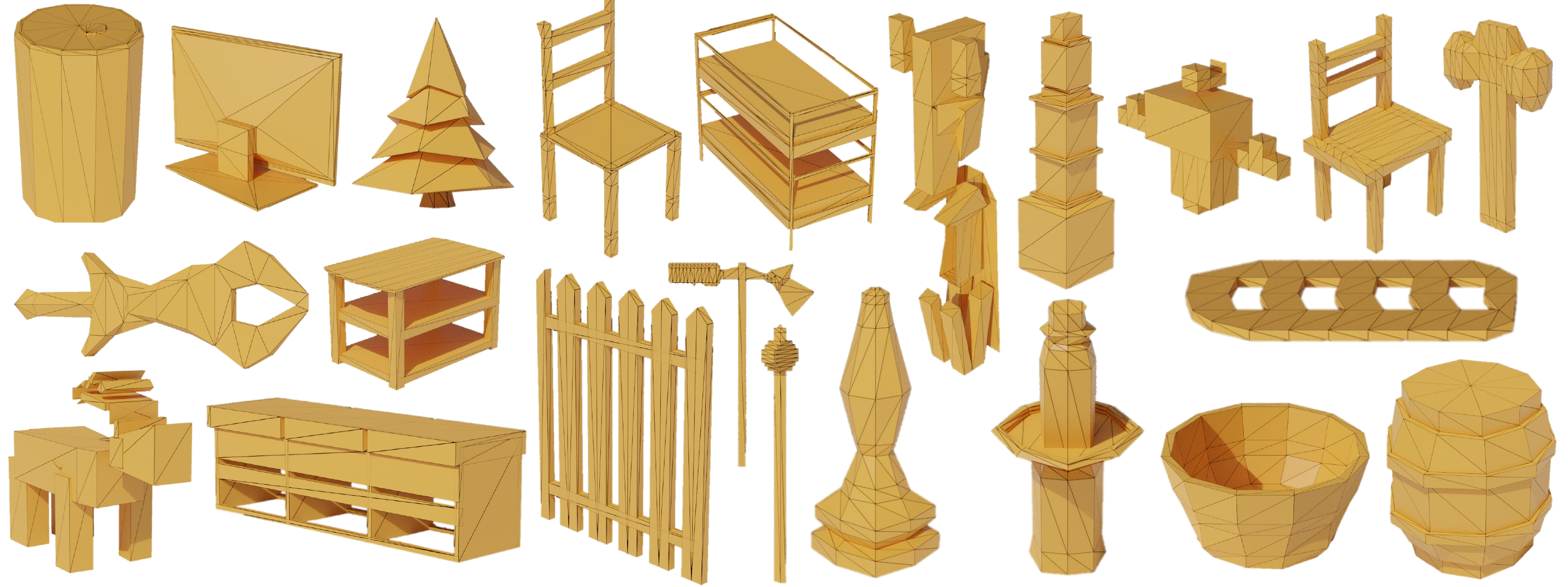}
      \caption{
        {\bf Gallery of generations from \ours.} We can generate high-quality and diverse meshes with artist-like created topology.
      }
      \label{fig:gallery}
      \vspace{-0.01\textheight}
    \end{figure*}

%% file: figs/tex/mesh_representation.tex
\begin{figure}
    \centering
    \includegraphics[width=1.0\linewidth]{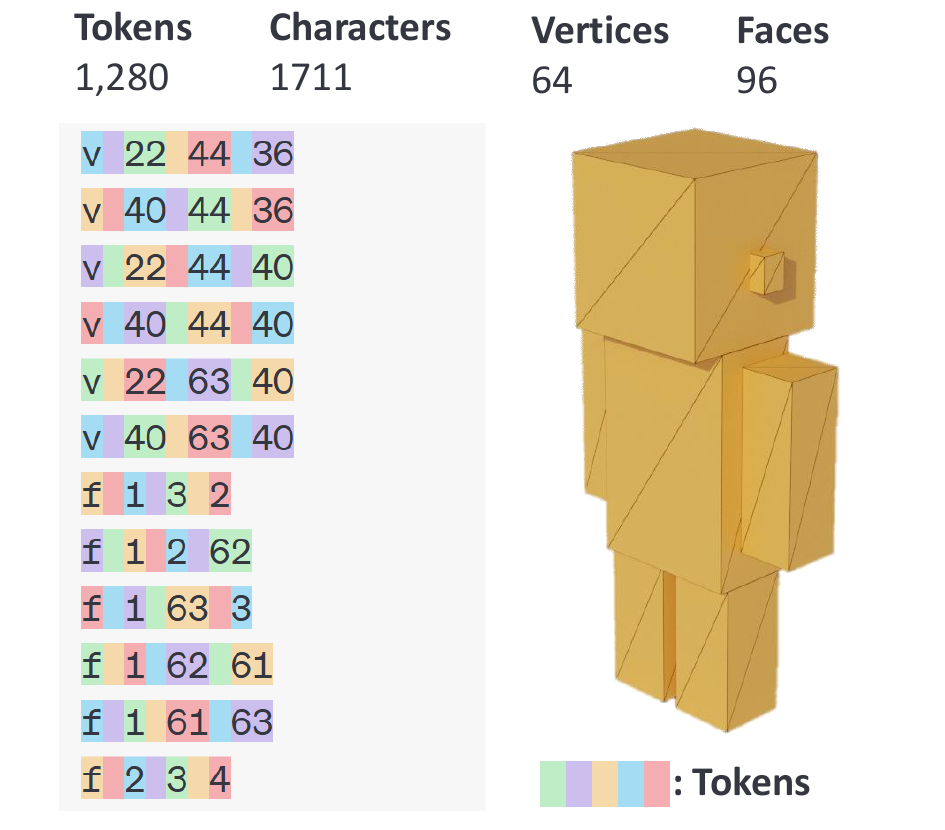}
    \caption{
        \textbf{Illustration of our 3D representation approach.} Left: A snippet of an OBJ file represented as plain text, containing vertex (\texttt{v}) and face (\texttt{f}) definitions. Right: The 3D object rendered from the OBJ file. We enable the LLM to process and generate 3D meshes by converting the mesh data into a textual format.
        }
    \label{fig:mesh_representation}
    \vspace{-0.02\textwidth}
\end{figure}

%% file: figs/tex/tokenizer2.tex
        \begin{figure}[t]
            \vspace{-0.03\textheight}
            \centering
            \includegraphics[width=\linewidth]{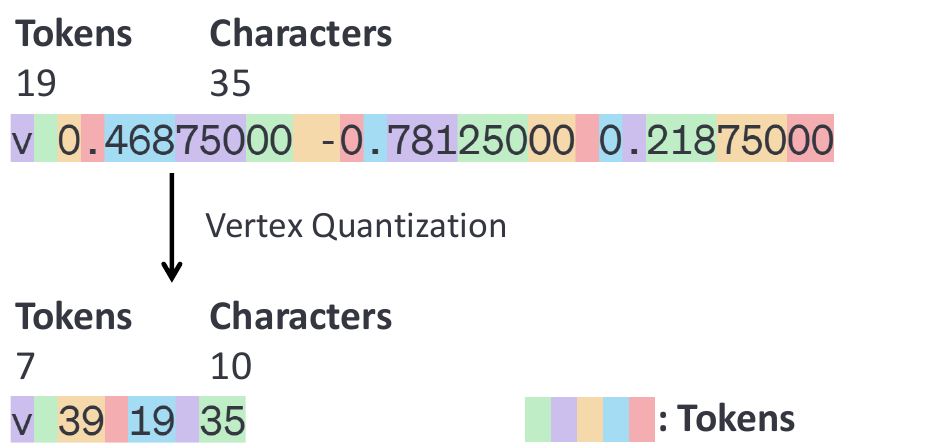}
            \caption{
                \textbf{Illustration of our vertex quantization method.} \emph{Top:} The original OBJ file represents vertex coordinates in decimal values, splitting a single coordinate into several tokens. \emph{Bottom:} After quantization, we represent the vertices as integers containing fewer tokens and are processed by LLM more efficiently.
                }
            \label{fig:tokenizer2}
            \vspace{-0.01\textheight}
        \end{figure}

%% file: figs/tex/dialog_pretrain_LLM.tex
    \begin{figure}[t]
        \centering
        \vspace{-0.03\textheight}
        \includegraphics[width=1.0\linewidth]{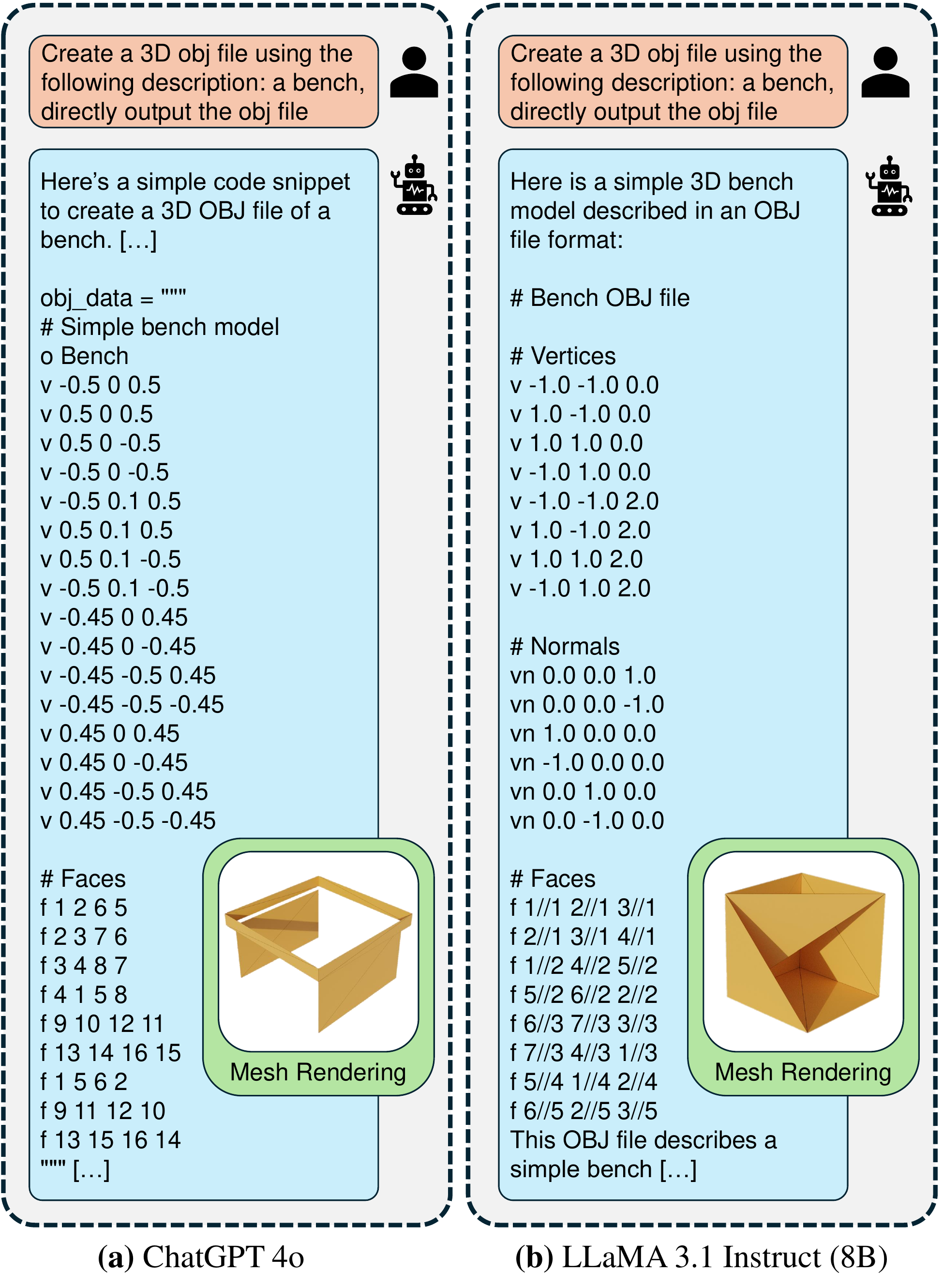}
        \vspace{-0.02\textheight}
        \caption{
            \textbf{Illustration of mesh generation capability from an LLM without finetuning.} \emph{Left:} results from ChatGPT-4o. \emph{Right:} results from LLaMA 3.1 8B-Instruct. Pretrained LLMs can generate simple 3D objects in text format; however, mesh quality and complexity are often unsatisfactory. OBJ files from the internet may vary slightly in format. The [...] indicates omitted text.
            }
        \label{fig:dialog_pretrain_LLM}
        \vspace{-0.01\textheight}
    \end{figure}

%% file: figs/tex/more_dialog_results.tex
\begin{figure*}
    \vspace{-0.03\textheight}
    \centering
        \includegraphics[width=1\linewidth,trim={0.0cm 0 0 0},clip]{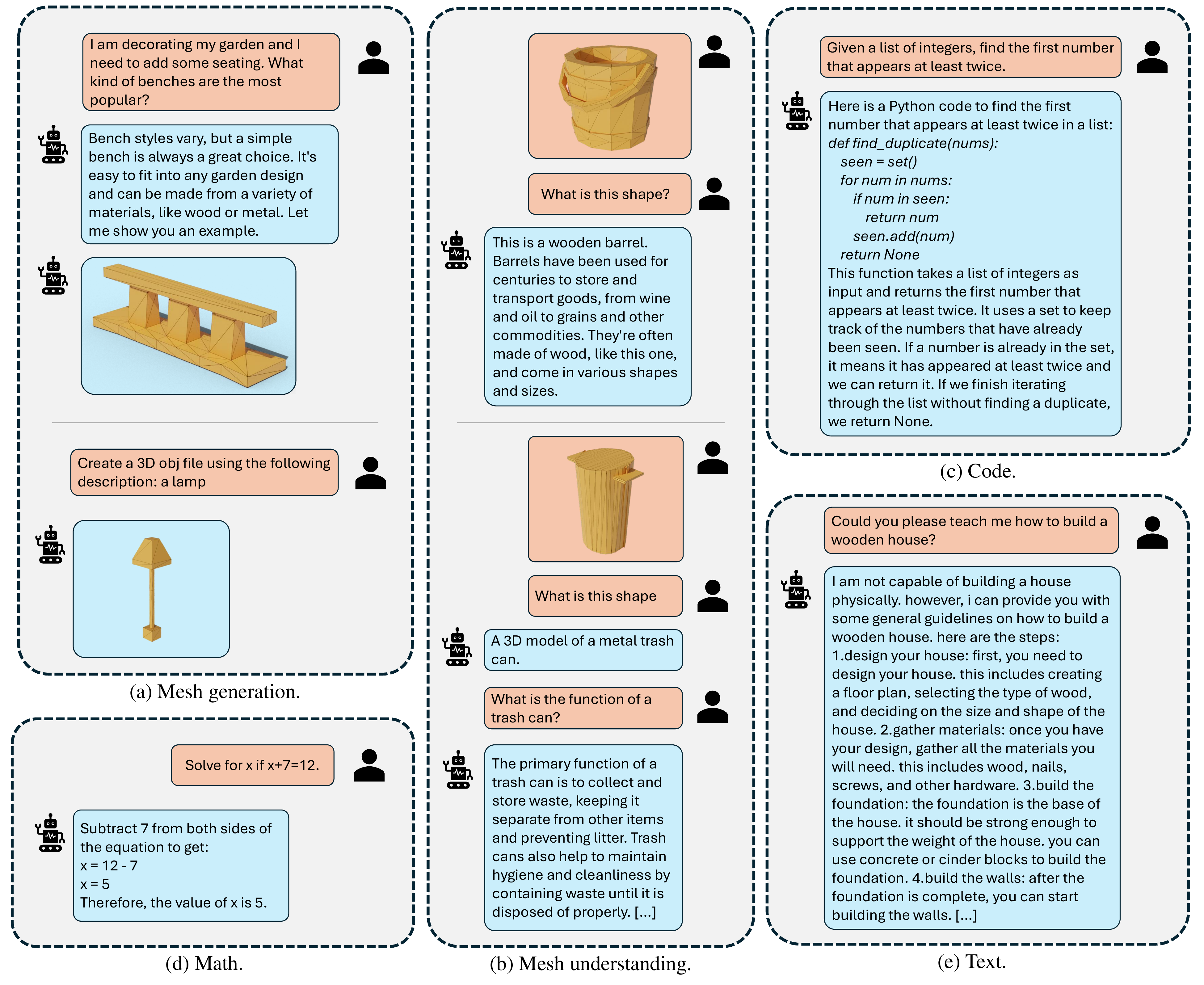}
    \vspace{-0.01\textheight}
    \caption{\textbf{More dialog results.} \ours achieves several new tasks, including mesh generation and understanding, while completing other tasks like the original LLM. [...]: we omit some text to make the snippet fit into the page.}
    \label{fig:more_dialog_results}
    \vspace{-0.00\textheight}
\end{figure*}

%% file: figs/tex/training_data.tex
\begin{figure*}
    \centering
    \vspace{-0.03\textheight}
    \includegraphics[width=\linewidth]{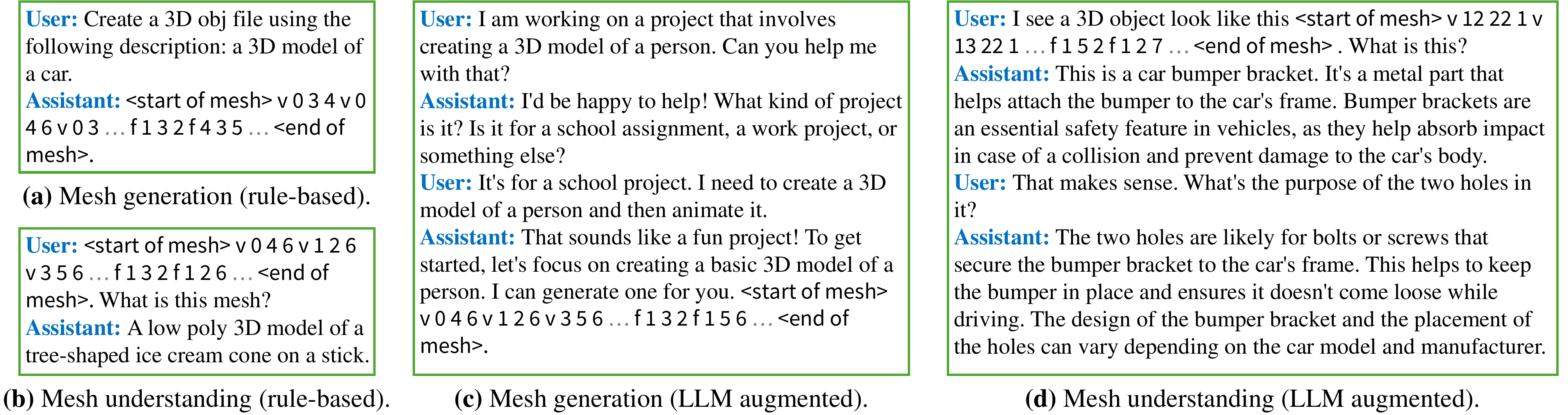}
    \caption{\textbf{Training dataset curated for \ours.} We use a combination of rule-based methods in \textbf{(a)} and \textbf{(b)} and LLM-augmented methods in \textbf{(c)} and \textbf{(d)} to construct an SFT dataset for mesh generation and understanding. \texttt{<start/end of mesh>} is shown here for illustration only and does not appear in the training data.}
    \label{fig:training_data}
\end{figure*}

%% file: figs/tex/train_loss.tex
\begin{figure}
    \centering
    \vspace{-0.02\textheight}
    \begin{tikzpicture}
        \node (img11) at (0,0) {\includegraphics[trim={1.3cm 1.0cm 0.0cm 1.8cm}, clip, width=0.9\linewidth]{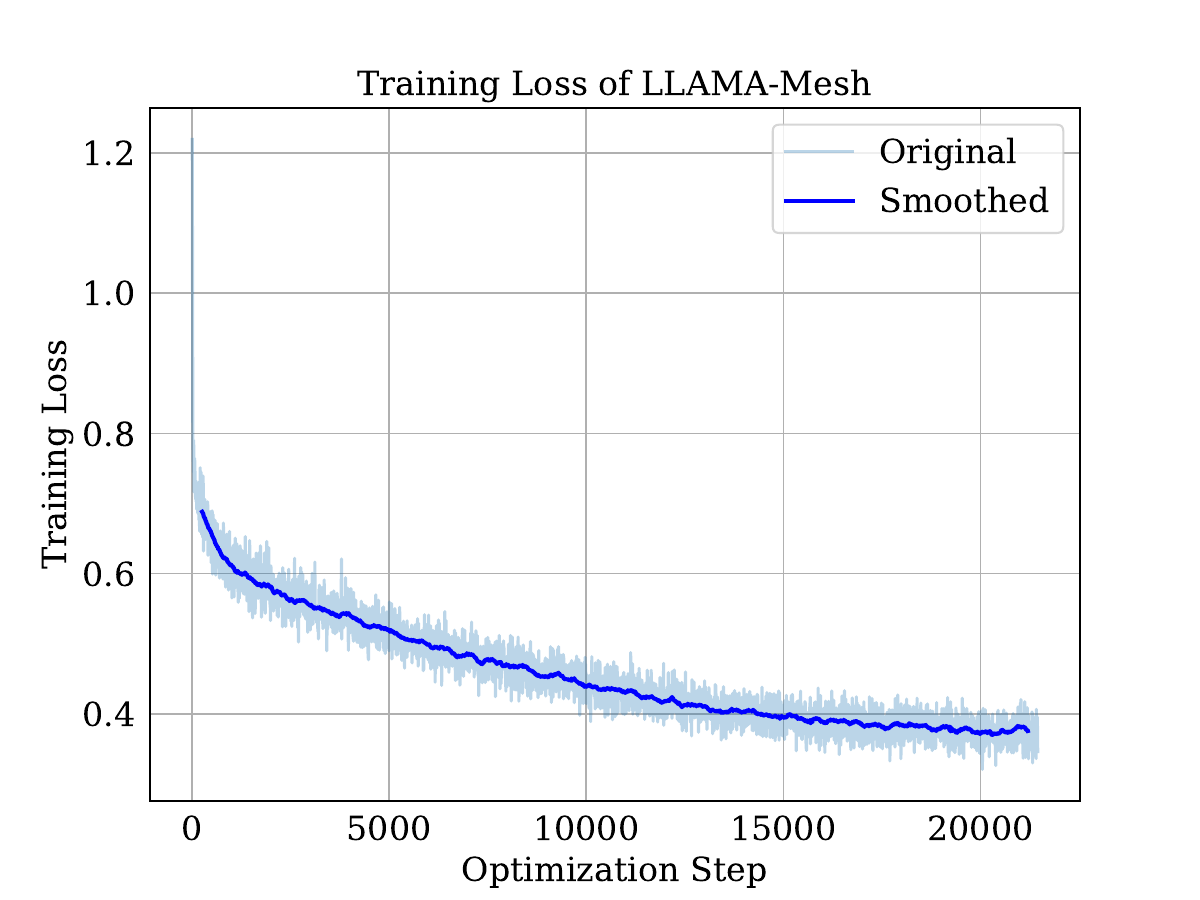}};
        \node[left=of img11, rotate=90, anchor=center, yshift=-.75cm] {\small{Training Loss}};
        \node[below=of img11, node distance=0cm, yshift=1.1cm, font=\color{black}] {\small{Optimization Step}};
    \end{tikzpicture}
    \vspace{-0.01\textheight}
    \caption{
        \textbf{Training loss of \ours.}
        The model adapts quickly to the new modality. We do not observe loss instabilities during training.
        Total training time comparisons are in Table~\ref{tab:training_time}.
    }
    \vspace{-0.00\textheight}
    \label{fig:train_loss}
\end{figure}

%% file: figs/tex/dataset_stat.tex
        \begin{table}
        	\vspace{-0.01\textheight}
        	\centering
        \begin{tabular}{l|lll}
            \toprule
            Dataset        & Items & \# Turns & Prop.\\
            \midrule
            Mesh Generation$^\dagger$& $125k$ & $8\times$ &  $40\%$\\
            Mesh Understanding$^\dagger$& $125k$ & $4\times$ & $20\%$ \\
            General Conversation~\cite{ding2023ultrachat} & $1$M & $1\times$ & $40\%$\\
            \bottomrule
        \end{tabular}
        	\caption{
                    \textbf{Dataset Statistics.} We list each dataset’s number of items, number of training turns per item, and the total sample proportions. Training is performed on a combined dataset, with each dataset resampled according to the ratio. We use a mix of mesh generation, mesh understanding, and general conversation data to equip LLMs with 3D capabilities while maintaining their language abilities. Datasets marked with $^\dagger$ are those we constructed.
                }
        	\label{tab:dataset_stat}
        \end{table}

%% file: figs/tex/mesh_gen_diversity.tex
            \begin{figure}
                \centering
                    \includegraphics[width=1\linewidth,trim={0.0cm 0 0 0},clip]{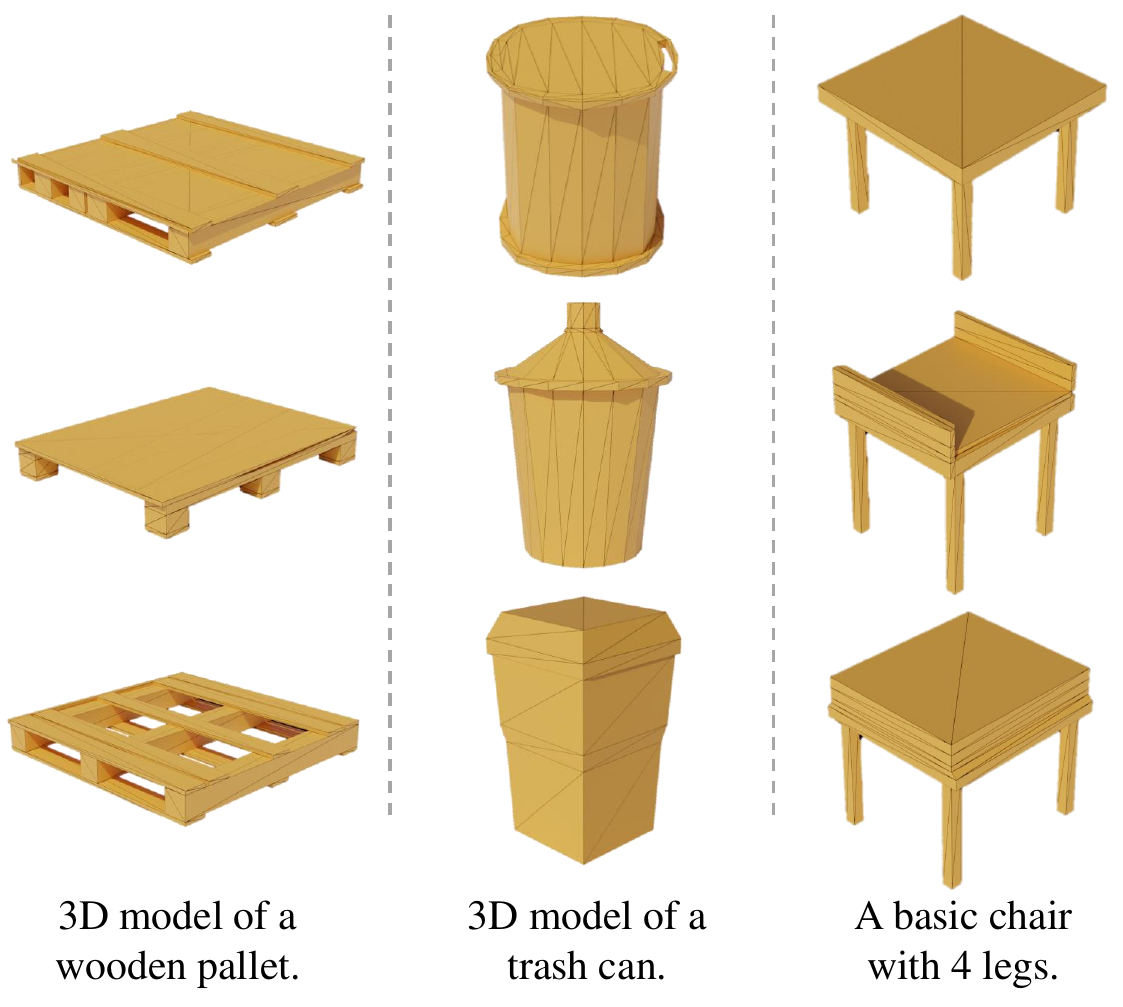}
                \caption{{\bf Diversity of generations.} \ours can generate diverse shapes given the same text prompt.}
                \label{fig:diversity}
            \end{figure}   

%% file: figs/tex/baseline_mesh_gen.tex
\begin{figure*}[th]
    \centering
    \vspace{-0.035\textheight}
    \includegraphics[width=0.99\linewidth]{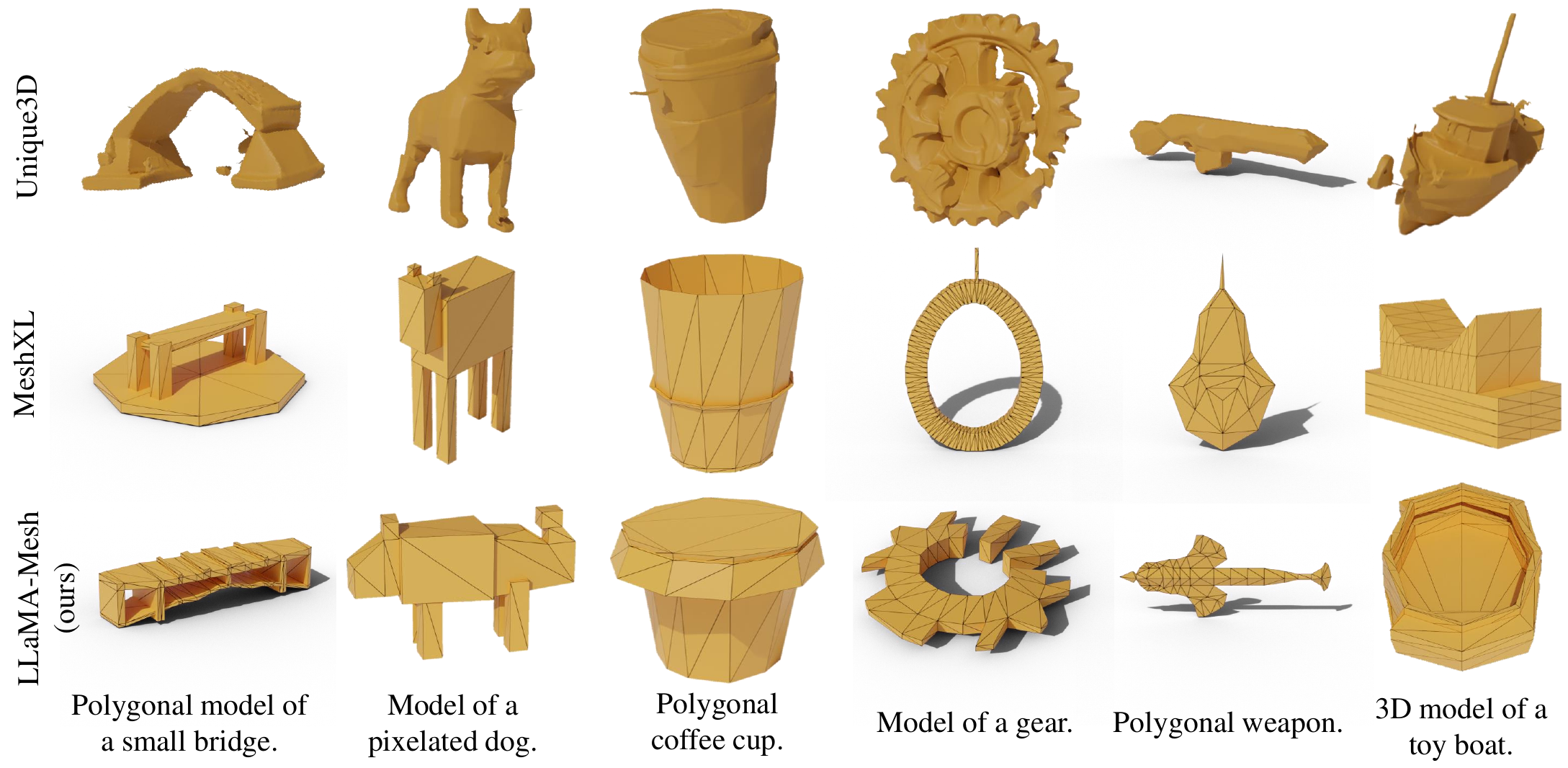}
    \vspace{-0.01\textheight}
    \caption{{\bf Comparison of \ours and baselines on text-to-mesh generation.} Our method achieves a competitive mesh quality.} 
    \label{fig:baseline_mesh_gen}
    \vspace{-0.00\textheight}
\end{figure*}

%% file: figs/tex/training_time.tex
        \begin{table}
	\centering
\begin{tabular}{l|ll>{\columncolor{cyan!20}}l}
    \toprule
    Method        & \multicolumn{2}{c}{MeshXL~\cite{chen2024meshxl}} & \ours\\
    \midrule
    Model Size    & $350$M & $1.3$B & $8$B \\
    GPU hours     & \num{6000} & \num{23232} & \num{2400} \\
    \bottomrule
\end{tabular}
        \vspace{-0.005\textheight}
	\caption{
            \textbf{Training time comparison.} Compared to MeshXL~\cite{chen2024meshxl}, \ours uses far fewer GPU hours despite its larger model size, benefiting from using pretrained LLM weights.
        }
	\label{tab:training_time}
    \vspace{-0.02\textheight}
\end{table}

%% file: figs/tex/ablation_llm_benchmark.tex
\begin{table*}[tb]
	\centering
	\begin{tabular}{l|l>{\columncolor{cyan!20}}lll}
		\toprule
		Metric        & LLaMA3.1 ($8$B)   & \ours ($8$B)    & LLaMA3.2 ($3$B) & LLaMA3.2 ($1$B)\\
		\midrule
		MMLU {\tiny(5-shot)}        & \num{66.07}     &   \num{61.74}     &\num{59.44}    &  \num{44.17}   \\
        PIQA {\tiny(0-shot)}&\num{81.01}&\num{79.16}&\num{75.52}&\num{74.10}\\
        Hellaswag {\tiny(0-shot)} &\num{79.19}&\num{77.35}&\num{70.47}&\num{60.80}\\
		GSM8K {\tiny(8-shot)}    & \num{77.18} & \num{62.09} & \num{66.94} & \num{34.27} \\
        
		\bottomrule
	\end{tabular}
	\caption{
            {\bf Does \ours preserve language capabilities?} We report the performance of \ours{} (8B) and compare it with base models of different sizes: LLaMA3.1 (8B), LLaMA3.2 (3B), and LLaMA3.2 (1B). The metrics include MMLU (5-shot), PIQA (0-shot), HellaSwag (0-shot), and GSM8K (8-shot), which assess the model's general knowledge, commonsense reasoning, and mathematical problem-solving abilities. \textbf{Takeaway:} Our method (in the \colorbox{cyan!20}{blue} column), after being fine-tuned to generate OBJ files, maintains language understanding and reasoning capabilities comparable to the base model while extending its functionality to 3D mesh generation.
        }
	\label{tab:ablation_llm_benchmark}
\end{table*}